\documentclass[pdflatex,sn-mathphys-num]{sn-jnl}

\usepackage{graphicx}%
\usepackage{multirow}%
\usepackage{amsmath,amssymb,amsfonts}%
\usepackage{amsthm}%
\usepackage{mathrsfs}%
\usepackage[title]{appendix}%
\usepackage{xcolor}%
\usepackage{textcomp}%
\usepackage{manyfoot}%
\usepackage{booktabs}%
\usepackage{algorithm}%
\usepackage{algorithmicx}%
\usepackage{algpseudocode}%
\usepackage{listings}%

\usepackage{ragged2e}
\usepackage{arydshln}
\usepackage{caption} 

\begin{document}

\title[Exploring the Influence of Dimensionality Reduction on Anomaly Detection Performance in Multivariate Time Series]{Exploring the Influence of Dimensionality Reduction on Anomaly Detection Performance in Multivariate Time Series}


\author*[1]{\fnm{Mahsun} \sur{Altin}}\email{altinma21@itu.edu.tr}

\author[2,3,4,5]{\fnm{Altan} \sur{Cakir}}\email{altan.cakir@itu.edu.tr}

\affil*[1]{\orgdiv{Department of Computer Engineering}, \orgname{Istanbul Technical University}, \orgaddress{\street{Sariyer}, \city{Istanbul}, \postcode{34467}, \country{Turkey}}}

\affil[2]{\orgdiv{Department of Physics Engineering}, \orgname{Istanbul Technical University}, \orgaddress{\street{Sariyer}, \city{Istanbul}, \postcode{34467}, \country{Turkey}}}

\affil[3]{\orgdiv{Artificial Intelligence, Data Science Research and Application Center}, \orgname{Istanbul Technical University}, \orgaddress{\street{Sariyer}, \city{Istanbul}, \postcode{34467}, \country{Turkey}}}

\affil[4]{\orgdiv{Parton Big Data Analytics and Consulting}, \orgname{ITU Ariteknokent}, \orgaddress{\street{Sariyer}, \city{Istanbul}, \postcode{34467}, \country{Turkey}}}

\affil[5]{\orgname{Adin.Ai}, \orgaddress{\city{Wilmington}, \state{Delaware}, \postcode{19806}, \country{United States}}}


\abstract{This paper presents an extensive empirical study on the integration of dimensionality reduction techniques with advanced unsupervised time series anomaly detection models, focusing on the MUTANT and Anomaly-Transformer models. The study involves a comprehensive evaluation across three different datasets: MSL, SMAP, and SWaT. Each dataset poses unique challenges, allowing for a robust assessment of the models' capabilities in varied contexts. The dimensionality reduction techniques examined include PCA, UMAP, Random Projection, and t-SNE, each offering distinct advantages in simplifying high-dimensional data. Our findings reveal that dimensionality reduction not only aids in reducing computational complexity but also significantly enhances anomaly detection performance in certain scenarios. Moreover, a remarkable reduction in training times was observed, with reductions by approximately 300\% and 650\% when dimensionality was halved and minimized to the lowest dimensions, respectively. This efficiency gain underscores the dual benefit of dimensionality reduction in both performance enhancement and operational efficiency. The MUTANT model exhibits notable adaptability, especially with UMAP reduction, while the Anomaly-Transformer demonstrates versatility across various reduction techniques. These insights provide a deeper understanding of the synergistic effects of dimensionality reduction and anomaly detection, contributing valuable perspectives to the field of time series analysis. The study underscores the importance of selecting appropriate dimensionality reduction strategies based on specific model requirements and dataset characteristics, paving the way for more efficient, accurate, and scalable solutions in anomaly detection.}




\maketitle

\section{Introduction \& Related Works}\label{sec1}

Real-world systems, encompassing a diverse array of domains from industrial equipment to healthcare apparatus \cite{b6}, \cite{b7}  and intricate financial networks \cite{b8}, are characterized by their continuous operation, incessantly producing extensive sequences of measurements through multi-sensor monitoring. Within this framework, the role of anomaly detection in multivariate time series data transcends a mere technical endeavor, establishing itself as an indispensable element for upholding system integrity and operational efficiency. The task of identifying malfunctions in large-scale monitoring data, primarily revolving around the detection of atypical temporal data points, is paramount in safeguarding against security breaches and mitigating substantial financial losses. However, the inherent rarity of anomalies, often obscured amidst voluminous normal data, poses significant challenges to data labeling, rendering it a laborious and costly endeavor. Consequently, this research is centered on the domain of unsupervised time series anomaly detection, a realm where the extraction of meaningful patterns from complex temporal dynamics is intricately linked with the challenge of detecting elusive anomalies in the absence of pre-labeled data.

In this context, time series analysis is integral to executing crucial functions, ranging from forecasting future variable trends — exemplified by stock market predictions — to the identification of anomalies within sequential data. The fundamental aim of time series anomaly detection methods is the identification of aberrant patterns within temporal data streams, a task of considerable significance across various industries. Notably, in the financial sector, the detection of anomalies plays a pivotal role in uncovering trading irregularities, potentially indicative of fraudulent activities or unexpected market movements. The detection process is further complicated by the infrequent nature of anomalies, typically overshadowed by an abundance of normative data. This scenario accentuates the demand for sophisticated anomaly detection methodologies, especially within unsupervised learning frameworks, where models are required to autonomously distinguish these infrequent anomalous events. In the realm of Multivariate Time Series Anomaly Detection, models are developed to analyze and interpret complex datasets that contain multiple interdependent variables over time \cite{b2}, \cite{b3}, \cite{b9}, \cite{b10}, \cite{b11}, \cite{b12}, \cite{b13}, \cite{b14}, \cite{b15}, \cite{b4}, \cite{b18}, \cite{b19}, \cite{b20}.  These models leverage advanced algorithms to identify patterns and relationships within the data, enabling them to detect anomalies that might be indicative of critical events or system malfunctions. Employing such models in unsupervised settings, where explicit labels for anomalies are absent, poses a unique challenge. These models must be adept at learning normal behavior patterns and identifying deviations without prior knowledge, relying solely on the intrinsic properties of the data. Such advancements are imperative for enhancing the efficacy and efficiency of anomaly detection in multifaceted, data-rich environments.

The landscape of anomaly detection models has undergone significant evolution in recent years, driven by the increasing complexity and volume of data across various domains. The current state of the art in this field is characterized by a trend towards developing more intricate and sophisticated models. These models are designed to capture the nuanced patterns and subtle irregularities indicative of anomalies in multivariate time series data. However, this advancement is not without its challenges and limitations. Dimension reduction is emerging as a significant technique whose impact on model performance requires extensive investigation. Dimensionality reduction methods like Principal Component Analysis (PCA), Uniform Manifold Approximation and Projection (UMAP), t-Distributed Stochastic Neighbor Embedding (t-SNE), and Random Projection offer a promising approach to simplifying the data without losing significant informational content. By reducing the number of variables under consideration, these techniques can help mitigate the issues associated with model complexity and long training times. This is especially pertinent in the context of complex datasets such as Soil Moisture Active Passive (SMAP), Mars Science Laboratory (MSL) and The Secure Water Treatment (SWaT), which are characterized by their high-dimensional nature.

The potential of dimensionality reduction to enhance the efficiency and effectiveness of anomaly detection models in handling such high-dimensional datasets is a critical area of study. Recent studies \cite{b4}, \cite{b5} holds particular significance in advancing the understanding of dimensionality reduction's role in data analysis, especially in complex datasets where both the reduction of dimensions and accurate outlier detection are pivotal. It holds the promise not only of reducing computational demands but also of maintaining, or even improving, model performance. This exploration is crucial in advancing the field of anomaly detection, particularly in light of the escalating demands for rapid and accurate detection in increasingly data-intensive environments. 

The remainder of this paper is orginized as follows, each focusing on a distinct aspect of anomaly detection in multivariate time series data through dimensionality reduction techniques. Section II provides an in-depth overview of standard dimensionality reduction methods, including PCA, UMAP, t-SNE, and Random Projection, highlighting their relevance in handling high-dimensional datasets. Section III explores various multivariate time series anomaly detection models, particularly emphasizing the methodologies and effectiveness of models like MUTANT \cite{b3} and Anomaly-Transformer \cite{b2}. Section IV details our empirical study, presenting the results of applying these dimensionality reduction techniques to real-world datasets, such as SMAP, MSL and SWaT, to evaluate the performance of the anomaly detection models. The discussion in Section V interprets these findings, placing them within the broader context of anomaly detection research. The paper concludes with Section VI, summarizing the key findings, discussing the implications for future research, and suggesting potential pathways for further exploration in the field of anomaly detection.

\section{Overview of Different Dimension Reduction Methods}\label{sec2}

This section provides a comprehensive overview of various dimension reduction methods, pivotal in the field of anomaly detection in multivariate time series data. Dimension reduction techniques are essential for simplifying complex datasets, thereby enhancing the performance and interpretability of anomaly detection models. We focus on four prominent methods: Principal Component Analysis (PCA), Uniform Manifold Approximation and Projection (UMAP), t-Distributed Stochastic Neighbor Embedding (t-SNE), and Random Projection. Each of these techniques offers a unique approach to reducing the dimensionality of data, balancing the retention of significant information with the reduction of computational complexity. In the following subsections, we will delve into the theoretical foundations of each method, discuss their advantages and limitations, and explore their applicability in the context of time series anomaly detection, setting the stage for their empirical evaluation in later sections of the paper.

\subsection{Principal Component Analysis (PCA)}\label{subsec1}

Principal Component Analysis (PCA) \cite{b17} is a statistical procedure that utilizes an orthogonal transformation to convert a set of possibly correlated variables into a set of linearly uncorrelated variables known as principal components. This technique is pivotal in the realm of dimensionality reduction, particularly in fields dealing with large and complex datasets. The primary goal of PCA is to capture the maximum variance present in the dataset with the fewest number of principal components. It achieves this by constructing a new coordinate system where the first axis corresponds to the direction of maximum variance, the second axis to the direction of the second most variance, and so forth.

The process of PCA begins with the computation of the covariance matrix, followed by the extraction of its eigenvalues and eigenvectors. These eigenvectors form the new axes for the transformed data space, and the eigenvalues indicate the amount of variance captured by each principal component. The transformation is defined mathematically as:

\begin{equation}
\Sigma = \frac{1}{m-1} \sum_{i=1}^{m} (x_i - \mu)(x_i - \mu)^T 
\label{pca}
\end{equation}

where \( \Sigma \) is the covariance matrix, \( x_i \) is a data point, \( \mu \) is the mean vector, and \( m \) is the number of data points. The data is then projected onto these new axes, resulting in a lower-dimensional representation of the original dataset.

In the context of anomaly detection, PCA's ability to emphasize variance and highlight patterns makes it a valuable tool. It simplifies the detection of outliers and anomalies, particularly in complex datasets like SMAP, MSL and SWaT. By reducing the number of dimensions, PCA not only helps in visualizing the data but also improves the computational efficiency of subsequent analyses. However, it's important to note that PCA assumes linear relationships between variables and may not perform well with non-linear data distributions. The choice of how many principal components to retain is often subjective and can impact the effectiveness of the anomaly detection.

In conclusion, PCA serves as a fundamental approach to dimensionality reduction, offering a balance between data simplification and the preservation of relevant information. Its application in the field of time series anomaly detection, especially in unsupervised learning scenarios, forms a crucial step in enhancing model performance and managing the challenges posed by high-dimensional data.

\subsection{Uniform Manifold Approximation and Projection (UMAP)}\label{subsec2}

Uniform Manifold Approximation and Projection (UMAP) \cite{b21} is a recent dimensionality reduction technique notable for effectively preserving both the local and global structure of data. This technique is grounded in manifold learning, which posits that high-dimensional data lies on a lower-dimensional manifold. UMAP has become favored for its ability to adeptly manage non-linear data structures, a feature that is particularly advantageous for complex datasets.

The essence of UMAP's approach involves constructing a high-dimensional graph to represent data point relationships, followed by optimizing a low-dimensional graph to mirror this structure as closely as possible. The optimization is rooted in fuzzy set theory, where UMAP minimizes the cross-entropy between two fuzzy sets that represent neighborhood relations in both the original and reduced spaces. The optimization function central to UMAP is given by:

\begin{equation}
\begin{split}
C(Y) &= \sum_{i, j} v_{ij} \log\left(\frac{1}{1 + a||y_i - y_j||^{2b}}\right) \\
&\quad + (1 - v_{ij}) \log\left(1 - \frac{1}{1 + a||y_i - y_j||^{2b}}\right)
\end{split}
\label{umap}
\end{equation}

In this equation, \( v_{ij} \) denotes the probability of points \( i \) and \( j \) being neighbors in the high-dimensional space, with \( y_i \) and \( y_j \) as their counterparts in the low-dimensional space. The parameters \( a \) and \( b \) are learned from the data, guiding the balance between local and global data structures. This function allows UMAP to retain critical topological and geometric structures of the high-dimensional data in a lower-dimensional representation.

In anomaly detection contexts, such as with the SMAP, MSL and SWaT datasets, UMAP's strength lies in its ability to reveal clusters and structures indicative of normal and anomalous patterns, particularly in data with complex relationships. However, it is computationally intensive and sensitive to hyperparameter choices. Despite these limitations, UMAP is an invaluable tool for dimensionality reduction, especially in unsupervised learning scenarios where understanding the intrinsic structure of data is crucial for detecting subtle anomalies.

UMAP stands as a robust method in the dimensionality reduction landscape, particularly effective in scenarios where preserving the intricate data structure is critical. Its application in time series anomaly detection models represents a significant step forward in handling complex, high-dimensional datasets and enhancing anomaly detection capabilities.

\subsection{Random Projection}\label{subsec3}

Random Projection is a dimensionality reduction technique that is part of the broader family of techniques known as linear projections. Unlike more complex methods like PCA or t-SNE, Random Projection reduces dimensionality through a very simple, yet effective, mechanism. It relies on the mathematical concept of the Johnson-Lindenstrauss lemma \cite{b16}, which suggests that distances between points in a high-dimensional space can be preserved in a lower-dimensional space using random linear projections.

The key idea behind Random Projection is to project the data from a high-dimensional space onto a randomly selected subspace of a lower dimension. This is achieved by multiplying the original data by a random matrix \( R \). Mathematically, if \( X \) is the original data matrix and \( Y \) is the transformed lower-dimensional data, the transformation can be represented as:

\begin{equation}
Y = XR 
\label{random_projection}
\end{equation}

where \( R \) is a \( d \times k \) random matrix (with \( d \) being the original dimensionality and \( k \) the reduced dimensionality). The elements of \( R \) are usually drawn from a normal distribution or other distributions that fulfill the requirements of the Johnson-Lindenstrauss lemma.

The appeal of Random Projection lies in its simplicity and efficiency, particularly for very large datasets. It can be significantly faster than methods like PCA, and the time complexity does not grow as quickly with the increase in dimensions. This makes it an attractive option for preliminary data analysis and for cases where computational resources are limited.

In the context of anomaly detection, Random Projection can be particularly useful for reducing the dimensionality of large-scale time series data, thereby making subsequent analysis more manageable. However, it's important to note that, since the projection is random, there is a possibility that some important features might be lost in the transformation. This trade-off between efficiency and the risk of losing potentially important information is a critical consideration when employing Random Projection in practice.

In conclusion, Random Projection offers a quick and computationally efficient approach to dimensionality reduction, particularly suited for large datasets where traditional methods may be too resource-intensive. While it may not always preserve all nuances of the original data, its simplicity and speed make it a valuable tool in the data scientist's arsenal, especially in the early stages of exploratory data analysis.

\subsection{t-Distributed Stochastic Neighbor Embedding (t-SNE)}\label{subsec4}

t-Distributed Stochastic Neighbor Embedding (t-SNE), as introduced by Laurens van der Maaten and Geoffrey Hinton \cite{b22}, is a non-linear technique primarily employed for reducing the dimensionality of high-dimensional datasets. It is particularly renowned for its efficacy in visualizing complex data structures in a low-dimensional space, typically in two or three dimensions. This capability to project high-dimensional data into two or three dimensions while preserving local relationships makes t-SNE an invaluable tool for exploratory data analysis and anomaly detection.

The algorithm of t-SNE commences by calculating pairwise similarities in the high-dimensional space, subsequently mapping these points to a lower-dimensional space (2D or 3D) to maintain these similarities as closely as possible. This mapping process is pivotal as it allows for the creation of visually interpretable maps that showcase the inherent clustering and structural nuances of the data, which is particularly beneficial in revealing underlying patterns crucial for anomaly detection.

Mathematically, the similarity of a datapoint \( x_j \) to \( x_i \) is represented as a conditional probability \( p_{j|i} \), defined as:

\begin{equation}
p_{j|i} = \frac{\exp(-||x_i - x_j||^2 / 2\sigma_i^2)}{\sum_{k \neq i} \exp(-||x_i - x_k||^2 / 2\sigma_i^2)}
\label{t_sne}
\end{equation}

In the reduced space, a similar probability \( q_{j|i} \) is calculated using a Student's t-distribution. The optimization objective is to minimize the Kullback-Leibler divergence between the probability distributions \( P \) in the high-dimensional space and \( Q \) in the reduced space.

The capacity of t-SNE to reduce dimensions to 2 or 3 allows for intuitive and accessible data visualization, aiding significantly in understanding complex datasets. This feature is especially useful in anomaly detection where the identification of outliers within clusters can be visually assessed.

Despite its strengths, t-SNE is not devoid of limitations. The technique is computationally demanding, particularly with larger datasets. The results of t-SNE are also sensitive to the choice of perplexity parameter and the initial configuration of the points in the reduced space. Furthermore, t-SNE is inclined to prioritize the preservation of local structures, possibly at the cost of global structure fidelity, which might lead to misinterpretations regarding the relative distances in the reduced space.

In essence, t-SNE provides a powerful mechanism for the visualization and interpretation of complex, high-dimensional datasets, particularly effective when reducing dimensions to 2 or 3. However, its usage in anomaly detection and other data analysis applications necessitates careful tuning and consideration of its limitations and strengths.

\section{Overview of Multivariate Time Series Anomaly Detection Models}\label{sec3}

In this section, we delve into the domain of unsupervised time series anomaly detection models, a critical area in the field of data science, particularly for applications where labeled data is scarce or non-existent. Unsupervised anomaly detection involves identifying unusual patterns or outliers in time series data without the guidance of a pre-labeled training set. This approach is essential in scenarios where anomalies are rare or unpredictable, and labeling is impractical due to the sheer volume of data or the complexity of defining what constitutes an anomaly. The focus here will be on exploring state-of-the-art models that have demonstrated efficacy in detecting anomalies in multivariate time series data. Models such as MUTANT (Robust anomaly detection for multivariate time series through temporal GCNs and attention-based VAE) and Anomaly-Transformer, both recent innovations in this space, will be examined in detail. These models represent the forefront of research in unsupervised anomaly detection, leveraging advanced techniques like graph convolutional networks and attention mechanisms to enhance their detection capabilities. By exploring their architectures, methodologies, and applications, we aim to shed light on the current capabilities and limitations of unsupervised anomaly detection models, setting the stage for our empirical evaluation of these models in conjunction with dimensionality reduction techniques.

\subsection{MUTANT}\label{subsec1}

The \textit{MUTANT} model, detailed in \textit{"Robust anomaly detection for multivariate time series through temporal GCNs and attention-based VAE,"} \cite{b3} presents a sophisticated approach to unsupervised anomaly detection in multivariate time series data. This model adeptly integrates Graph Convolutional Networks (GCNs) and Variational Auto-Encoders (VAEs) with an attention mechanism, addressing the dual challenges of discerning time-varying correlations between variables and assessing the significance of these variables over different periods for effective anomaly detection.

At its core, the MUTANT model constructs feature graphs for each time window, employing GCNs to learn embeddings that reflect the correlations within time series data. This aspect of the model is encapsulated by the GCN layer-wise propagation rule:

\begin{equation}
H_{t}^{(l+1)} = \text{ReLU}\left(\tilde{D}_t^{-\frac{1}{2}} \tilde{A}_t \tilde{D}_t^{-\frac{1}{2}} H_{t}^{(l)} W_{t}^{(l)}\right)
\label{MUTANT_equation_2}
\end{equation}

Furthermore, an LSTM-based attention module is utilized to gauge the relevance of variables over time. This mechanism is defined by a series of equations governing the LSTM unit's functionality:

\begin{equation}
f_t = \sigma(W_f [h_{t-1}; \tilde{x}_t] + b_f)
\label{MUTANT_equation_3}
\end{equation}
\begin{equation}
i_t = \sigma(W_i [h_{t-1}; \tilde{x}_t] + b_i)
\label{MUTANT_equation_4}
\end{equation}
\begin{equation}
o_t = \sigma(W_o [h_{t-1}; \tilde{x}_t] + b_o)
\label{MUTANT_equation_5}
\end{equation}

The Pearson Correlation Coefficient plays a vital role in constructing these feature graphs:

\begin{equation}
\rho_{ij} = \frac{\text{Cov}(f_{ti}, f_{tj})}{\sqrt{\text{Var}[f_{ti}] \cdot \text{Var}[f_{tj}]}} 
\label{MUTANT_equation_1}
\end{equation}

The MUTANT model demonstrates significant strengths in anomaly detection, particularly in handling time-varying correlations and variable importance. However, its intricate architecture necessitates careful parameter tuning and consideration, especially in diverse or noisy datasets.

In our analysis of the MUTANT model, a critical aspect to consider is the model's inherent structure with respect to input dimensionality. The model's design imposes a lower limit on the dimensionality of the input dataset it can process, specifically requiring that the input data have no fewer than 8 dimensions. This constraint is attributed to the architectural intricacies of the model, which necessitate a minimum threshold of input features for effective functioning. Consequently, in scenarios involving dimensionality reduction, our empirical evaluations had to ensure that the resulting datasets post-reduction did not fall below this minimum dimensionality of 8. This precaution allowed us to maintain compatibility with the MUTANT model's default structure and effectively evaluate its performance on datasets with reduced dimensions while adhering to the model's operational constraints. The necessity to accommodate this minimum dimensionality requirement is an important consideration, particularly when dealing with inherently low-dimensional data.

\subsection{Anomaly-Transformer}\label{subsec2}

The \textit{Anomaly-Transformer} model, as delineated in the paper \textit{"Anomaly Transformer: Time Series Anomaly Detection with Association Discrepancy,"} \cite{b2} represents an innovative approach in the domain of unsupervised time series anomaly detection. The model's core concept revolves around the utilization of \textit{association discrepancy} to discern anomalies.

\subsubsection{Key Components and Equations}\label{subsubsec1}

\begin{enumerate}
    \item \textbf{Multi-Level Quantification:} The model enhances the final results by averaging the association discrepancy across multiple layers. This approach, leveraging multi-layer design, has been empirically validated to enhance the effectiveness of anomaly detection.
    
    \item \textbf{Association Discrepancy Calculation (Algorithm 2):} The model employs a novel algorithm to calculate the association discrepancy:
        \begin{equation}
            P = \text{Mean}(P_{\text{dim=1}}), \; P \in \mathbb{R}^{L \times N \times N}
        \label{ANOMALY_TRANSFORMER_equation_1}
        \end{equation}
        \begin{equation}
            S = \text{Mean}(S_{\text{dim=1}}), \; S \in \mathbb{R}^{L \times N \times N}
        \label{ANOMALY_TRANSFORMER_equation_2}
        \end{equation}
        \begin{equation}
            R = \text{KL}(P \,||\, S)_{\text{dim=-1}} + \text{KL}(S \,||\, P)_{\text{dim=-1}}, \; R \in \mathbb{R}^{L \times N}
        \label{ANOMALY_TRANSFORMER_equation_3}
        \end{equation}
        \begin{equation}
            R = \text{Mean}(R, \text{dim=0}), \; R \in \mathbb{R}^{N \times 1}
        \label{ANOMALY_TRANSFORMER_equation_4}
        \end{equation}
    
    \item \textbf{Association-Based Criterion (Algorithm 3):} This algorithm is pivotal for computing the anomaly score for each time point:
        \begin{equation}
            CAD = \text{Softmax}(-\text{AssDis}(P_{S}; X)_{\text{dim=0}}), \; CAD \in \mathbb{R}^{N \times 1}
        \label{ANOMALY_TRANSFORMER_equation_5}
        \end{equation}
        \begin{equation}
            C_{\text{Recon}} = \text{Mean}((X - \hat{X})^2, \text{dim=1}), \; C_{\text{Recon}} \in \mathbb{R}^{N \times 1}
        \label{ANOMALY_TRANSFORMER_equation_6}
        \end{equation}
        \begin{equation}
            C = CAD \times C_{\text{Recon}}, \; C \in \mathbb{R}^{N \times 1}
        \label{ANOMALY_TRANSFORMER_equation_7}
        \end{equation}
    
    \item \textbf{Anomaly Score Calculation:} The Anomaly Score is determined using the multiplication method, which combines association discrepancy and reconstruction error:
        \begin{equation}
            \begin{split}
                \text{AnomalyScore}(X) &= \text{Softmax}(-\text{AssDis}(P_{S}; X)) \\
                &\quad \times (X_{i} - \hat{X}_{i})^2, \; \text{for } i=1 \ldots N
            \end{split}
            \label{ANOMALY_TRANSFORMER_equation_8}
        \end{equation}

    \item \textbf{Optimization Strategy:} The model adopts a minimax strategy for optimization, focusing on the reconstruction loss and the association discrepancy.
\end{enumerate}

The \textit{Anomaly-Transformer} model, through its unique focus on association discrepancy, is adept at capturing complex temporal relationships, rendering it highly effective for datasets with subtle and nuanced anomalies. Nonetheless, the complexity of the model necessitates meticulous parameter tuning, underscoring the importance of precision in its deployment.

\section{Empirical Study}\label{sec4}

In this section, we embark on an empirical study to rigorously evaluate the performance of the discussed dimensionality reduction techniques and unsupervised time series anomaly detection models, specifically focusing on the MUTANT and Anomaly-Transformer models. This investigation is conducted using the SMAP, MSL and SWaT datasets, which are characterized by their complex and multivariate nature. The primary aim is to assess how effectively these models, in conjunction with the dimensionality reduction methods of PCA, UMAP, t-SNE, and Random Projection, can identify anomalies in such challenging datasets. We systematically apply each dimensionality reduction technique with different numbers to the datasets and then deploy the anomaly detection models to examine their performance in terms of accuracy, computational efficiency, and their ability to handle the intricacies of high-dimensional time series data. The results of these experiments will provide valuable insights into the strengths and limitations of each approach, offering a comprehensive understanding of their applicability in real-world scenarios. This empirical analysis is crucial for establishing benchmarks in the field and for guiding future research directions in the realm of time series anomaly detection.

\subsection{Proposed Model Architecture}\label{subsec1}

\begin{figure*}
\centerline{\includegraphics[width=6in]{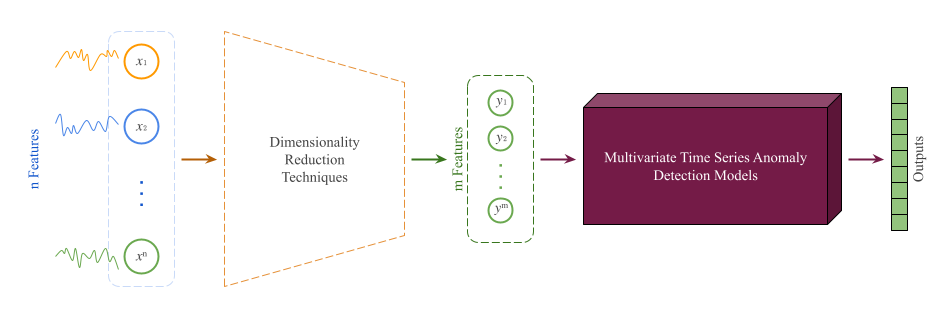}}
\caption{Detailed Schematic of the Proposed Architecture Incorporating Dimensionality Reduction: Transitioning from n Features to m Features}
\label{architecture}
\end{figure*}

Our proposed model structure for anomaly detection in multivariate time series data is a sequential integration of dimensionality reduction techniques followed by advanced anomaly detection models. This structure is visually represented in Figure \ref{architecture} and can be described as follows:

\begin{enumerate}
    \item \textbf{Initial Data Input (\( n \) Numbers)}: The model begins with an input of multivariate time series data, consisting of \( n \) dimensions or features. This data, which could be derived from various sources like sensors in the SMAP, MSL and SWaT datasets, is typically high-dimensional, potentially containing noise and redundant information.

    \item \textbf{Application of Dimensionality Reduction Techniques}: The next stage involves the application of dimensionality reduction techniques to the initial high-dimensional data. This step is crucial for simplifying the data structure, enhancing model efficiency, and potentially improving anomaly detection performance. We consider four major techniques in our framework:
    \begin{itemize}
        \item Principal Component Analysis (PCA)
        \item Uniform Manifold Approximation and Projection (UMAP)
        \item t-Distributed Stochastic Neighbor Embedding (t-SNE)
        \item Random Projection
    \end{itemize}
    Each of these techniques reduces the data from \( n \) dimensions to \( m \) dimensions, where \( m \) is significantly smaller than \( n \). The choice of technique and the degree of dimensionality reduction depend on the specific characteristics of the dataset and the requirements of the anomaly detection task.

    \item \textbf{Reduced Data Output (\( m \) Numbers)}: The output from the dimensionality reduction stage is a transformed dataset with \( m \) dimensions. This reduced dataset retains the most significant features of the original data, essential for effective anomaly detection, while minimizing information loss.

    \item \textbf{Application of Multivariate Time Series Anomaly Detection Models}: The final stage involves feeding the dimensionality-reduced data into sophisticated anomaly detection models. In our study, we focus on two models:
    \begin{itemize}
        \item MUTANT: A model that leverages temporal Graph Convolutional Networks (GCNs) and an attention-based Variational Auto-Encoder (VAE).
        \item Anomaly-Transformer: A model that uses association discrepancy for anomaly detection.
    \end{itemize}
    These models are designed to identify anomalies by analyzing the temporal and structural patterns in the reduced feature space. Their ability to detect anomalies in lower-dimensional data is a critical aspect of our empirical study.
\end{enumerate}

The proposed model structure aims to combine the strengths of dimensionality reduction techniques with advanced anomaly detection models to effectively identify anomalies in complex, multivariate time series data. This integrated approach is anticipated to not only enhance the accuracy of anomaly detection but also to improve computational efficiency, making it well-suited for handling large-scale datasets. The empirical evaluation of this model structure, as outlined in our study, will provide insights into its efficacy and potential for practical applications in various fields.

\subsection{Datasets}\label{subsec2}

\begin{table*}
\centering
\caption{Comprehensive Details of Utilized Datasets: MSL, SMAP, and SWaT with Training and Test Samples, Dimensions, and Anomaly Percentages}
\label{tab:dataset_details}
\begin{tabular}{cccccc} 
\toprule
Dataset & Train  & Test   & \#dimensions & Anomalies (\%) & Citation  \\ 
\midrule
MSL     & 58317  & 73729  & 55           & 10.72          & \cite{b1} \\
SMAP    & 135183 & 427617 & 25           & 13.13          & \cite{b1} \\
SWaT    & 495000 & 449919 & 51           & 11.98          & \cite{b23}, \cite{b24} \\
\bottomrule
\end{tabular}
\end{table*}

The datasets employed in this study, as detailed in Table \ref{tab:dataset_details}, are instrumental for understanding the nuances of anomaly detection in multivariate time series data. Each dataset presents unique challenges and characteristics, providing a diverse testing ground for our research.

\begin{enumerate}
    \item \textbf{The Mars Science Laboratory (MSL) Rover Dataset:} This dataset from NASA encompasses monitoring data with 55 metrics. As shown in Table \ref{tab:dataset_details}, the MSL dataset consists of 58,317 training and 73,729 test samples, with approximately 10.72\% of the test data containing anomalies. The dataset is significant for its real-world applicability in space exploration and the complexity of its variables.

    \item \textbf{The Soil Moisture Active Passive (SMAP) Satellite Dataset:} Also from NASA, the SMAP dataset includes data from 55 entities. With a considerably larger number of samples, 135,183 for training and 427,617 for testing, the dataset has about 13.13\% anomalies in the test set. The SMAP dataset provides a broad perspective on satellite data monitoring and anomaly detection.

    \item \textbf{The Secure Water Treatment (SWaT) Dataset:} Originating from an industrial water treatment plant, the SWaT dataset represents a scenario of continuous operations over 11 days, including both normal conditions and attack scenarios. The dataset, encompassing 495,000 training and 449,919 test samples, covers data from various sensors and actuators across 51 metrics. With 11.98\% anomalies in the test set, the SWaT dataset offers a unique insight into anomaly detection in industrial control systems.
\end{enumerate}

These datasets, each with their distinct characteristics and anomaly profiles, are crucial for evaluating the efficacy of dimensionality reduction techniques and time series anomaly detection models. Their variety ensures that our findings and conclusions are robust and applicable across different real-world scenarios.

\section{Discussion of Results}\label{sec5}

\begin{sidewaystable}
\caption{Extensive Performance Comparison of MUTANT and Anomaly-Transformer Models Under Various Dimensionality Reduction Techniques Across MSL, SMAP and SWaT Datasets}
\label{tab:performance}
\begin{tabular}{cccccccccccc} 
\multirow{2}{*}{\Centering{}\textbf{Model}} & \multirow{2}{*}{\Centering{}\begin{tabular}[c]{@{}c@{}}\textbf{\# Dimensions}\\\textbf{Remaining}\end{tabular}} & \multirow{2}{*}{\Centering{}\begin{tabular}[c]{@{}c@{}}\textbf{DR Layer}\\\textbf{(Technique)}\end{tabular}} & \multicolumn{3}{c}{\textbf{MSL}} & \multicolumn{3}{c}{\textbf{SMAP}} & \multicolumn{3}{c}{\textbf{SWaT}} \\ 
\cmidrule(lr){4-6}\cmidrule(lr){7-9}\cmidrule(lr){10-12}
& &  & \textbf{Precision} & \textbf{Recall} & \textbf{F1-score} & \textbf{Precision} & \textbf{Recall} & \textbf{F1-score}  & \textbf{Precision} & \textbf{Recall} & \textbf{F1-score}   \\ 
\hline\hline
\multirow{7}{*}{\Centering{}{\rotatebox[origin=c]{90}{MUTANT}}} & (Original) & None & \textbf{0.9464} & 0.9520 & 0.9492 & 0.9658 & 0.9787 & 0.9722 & 0.9805 & 0.9881 & 0.9842  \\ 
\cdashline{2-12}[1pt/1pt]
& \multirow{3}{*}{\Centering{}\begin{tabular}[c]{@{}c@{}}To Half Dim.\\27 - 12 - 25\end{tabular}} & PCA & 0.9307 & 0.9807 & 0.9551 & 0.9725 & 0.9630 & 0.9678 & 0.9729 & \textbf{1.0000} & \textbf{0.9863}     \\
& & Rand. Proj.  & 0.8619 & \textbf{1.0000} & 0.9258 & 0.9703 & 0.9782 & 0.9742 & 0.9782 & 0.9882 & 0.9832  \\
& & UMAP         & 0.8846 & 0.9762 & 0.9281 & 0.9836 & 0.9453 & 0.9640 & 0.9491 & 0.9838 & 0.9661  \\ 
\cdashline{2-12}[1pt/1pt]
& \multirow{3}{*}{\Centering{}\begin{tabular}[c]{@{}c@{}}To Lowest Dim.\\8 - 8 - 8\end{tabular}}  & PCA & 0.9184 & 0.9848 & 0.9505 & 0.9882 & 0.9659 & \textbf{0.9769} & 0.9632 & 0.9866 & 0.9748  \\
& & Rand. Proj.  & 0.9331 & 0.9762 & 0.9542   & 0.9550 & \textbf{0.9866} & 0.9706 & 0.9728 & 0.9856 & 0.9792  \\
& & UMAP         & 0.9341 & 0.9914 & \textbf{0.9619} & \textbf{0.9913} & 0.9399 & 0.9649 & \textbf{0.9833} & 0.9788 & 0.9810  \\ 
\hline\hline
\multirow{12}{*}{\Centering{}{\rotatebox[origin=c]{90}{Anomaly-Transformer}}} & (Original) & None & 0.9188 & 0.9473 & 0.9329   & 0.9381 & 0.9939 & 0.9652 & 0.8844 & \textbf{1.0000} & 0.9386  \\ 
\cdashline{2-12}[1pt/1pt]
& \multirow{3}{*}{\Centering{}\begin{tabular}[c]{@{}c@{}}To Half Dim. \\27 - 12 - 25\end{tabular}} & PCA & 0.9146 & 0.9436 & 0.9289 & 0.9111 & 0.9916 & 0.9497 & 0.9223 & \textbf{1.0000} & 0.9596  \\
& & Rand. Proj.  & 0.9191 & 0.9773 & 0.9473 & 0.9160 & 0.9950 & 0.9539 & 0.8889 & \textbf{1.0000} & 0.9412  \\
& & UMAP         & 0.9178 & 0.9735 & 0.9448 & 0.9264 & \textbf{0.9993} & 0.9615 & 0.8482 & \textbf{1.0000} & 0.9179  \\
\cdashline{2-12}[1pt/1pt]
& \multirow{4}{*}{\Centering{}3 - 3 - 3} & PCA & 0.9172 & 0.9676 & 0.9417   & 0.9072 & 0.9945 & 0.9489 & 0.9706 & 0.9495 & \textbf{0.9600}     \\
& & Rand. Proj.  & 0.9180 & \textbf{0.9793} & \textbf{0.9477}   & 0.9335 & 0.9919 & 0.9618 & \textbf{0.9891}    & 0.8619 & 0.9212  \\
& & UMAP         & 0.9171 & 0.9560 & 0.9361 & 0.9320 & 0.9915 & 0.9608 & 0.9807 & 0.9229 & 0.9509  \\
& & t-SNE        & 0.9164 & 0.9490 & 0.9324 & 0.9310 & 0.9962 & 0.9625 & 0.9843 & 0.9082 & 0.9447  \\ 
\cdashline{2-12}[1pt/1pt]
& \multirow{4}{*}{\Centering{}\begin{tabular}[c]{@{}c@{}}To Lowest Dim.\\2 - 2 - 2\end{tabular}}  & PCA & 0.9180 & 0.9683 & 0.9425   & 0.9070 & 0.9930 & 0.9481 & 0.9492 & 0.9696 & 0.9593  \\
& & Rand. Proj.  & \textbf{0.9210} & 0.9473 & 0.9340   & \textbf{0.9429} & 0.9524 & 0.9476 & 0.9876 & 0.8862 & 0.9341  \\
& & UMAP         & 0.9183 & 0.9677 & 0.9423 & 0.9330 & 0.9945 & 0.9628 & 0.9890 & 0.8871 & 0.9352  \\
& & t-SNE        & 0.9197 & 0.9749 & 0.9465 & 0.9353 & 0.9977 & \textbf{0.9655} & 0.9854 & 0.9237 & 0.9536  \\ 
\end{tabular}
\end{sidewaystable}

The comprehensive analysis of the performance of the MUTANT and Anomaly-Transformer models under various dimensionality reduction scenarios across the MSL, SMAP, and SWaT datasets has yielded insightful findings. These results, as detailed in Table \ref{tab:performance}, not only shed light on the individual efficacy of each model and dimensionality reduction technique but also reveal how they interact with different types of datasets. In this discussion, we delve deeper into these interactions, examining the performance nuances of each model with respect to original data, the impact of dimensionality reduction, and their overall performance across various datasets. Additionally, we explore how each dimensionality reduction technique influences the models' abilities to detect anomalies and how the specific characteristics of each dataset play a role in this dynamic. This multi-faceted analysis aims to provide a comprehensive understanding of the strengths and limitations of these advanced anomaly detection methods in the context of multivariate time series data.

\subsection{Models}\label{subsec1}

In the realm of unsupervised anomaly detection, the performance of models is heavily influenced by the nature of the data and the techniques used for preprocessing. This section focuses on dissecting the performance of two cutting-edge models: MUTANT and Anomaly-Transformer. Each model's unique approach to handling multivariate time series data under different dimensionality reduction scenarios is critically analyzed. We delve into how these models perform with their original data structure, the impact of dimensionality reduction on their efficacy, and the overall performance in varied dataset contexts. This comprehensive examination aims to reveal the strengths and limitations of each model, providing insights into their optimal application scenarios.

\subsubsection{MUTANT}\label{subsubsec1}

The MUTANT model, known for its robust approach to anomaly detection in multivariate time series data, is a focal point of our analysis. This model, leveraging the synergy of Graph Convolutional Networks and attention-based Variational Auto-Encoders, is designed to capture intricate time-dependent correlations and variable importance. In this subsection, we explore the MUTANT model's performance across different datasets and dimensionality reduction techniques. Starting with its original data performance, we assess how the model fares with unaltered data inputs, setting a benchmark for its capability in handling high-dimensional data. This exploration forms the basis for further evaluating the impact of dimensionality reduction techniques on the model's anomaly detection proficiency.

\paragraph{Original Data Performance}
In its original configuration, the MUTANT model demonstrates a strong ability to process high-dimensional data, as evidenced by its high precision on the MSL dataset. This performance is indicative of the model's inherent capabilities in handling complex, multivariate datasets without the need for dimensionality reduction. However, when observing the F1-score, the model exhibits a remarkable performance on the dimensionally reduced SMAP dataset, suggesting that certain reductions can potentially enhance its anomaly detection capabilities.

\paragraph{Impact of Dimensionality Reduction}
The application of dimensionality reduction techniques does not uniformly affect the MUTANT model's performance; in fact, certain techniques like UMAP, when applied to reduce the dataset to the lowest dimensionality (8 features), significantly improve the model's F1-score, particularly for the SWaT dataset. This improvement indicates that strategic dimensionality reduction can aid the model in focusing on the most relevant features for anomaly detection, thereby enhancing its effectiveness.

\paragraph{Best Overall Performance}
The MUTANT model's best overall performance, especially in terms of F1-score, is observed when UMAP is used to reduce the dimensionality to its minimum viable threshold for the SWaT dataset. This finding underscores the importance of selecting an appropriate dimensionality reduction technique that aligns with the model's structural requirements and the specific characteristics of the dataset being analyzed.

\subsubsection{Anomaly-Transformer}\label{subsubsec2}

The Anomaly-Transformer model represents a novel approach in the landscape of time series anomaly detection. Characterized by its utilization of association discrepancies for anomaly identification, this model has shown promise in effectively discerning irregular patterns within multivariate time series data. In this subsection, we aim to unravel the performance dynamics of the Anomaly-Transformer model under various settings. Initial focus is given to its performance with the original, unaltered data. This analysis is crucial to understand the model's baseline capabilities before the application of dimensionality reduction techniques, providing a clear perspective on its native efficiency and accuracy in handling complex datasets.

\paragraph{Original Data Performance}
The Anomaly-Transformer model shows its prowess in the original, high-dimensional setting of the SWaT dataset, achieving the highest F1-score among all scenarios tested. This performance is indicative of the model's robustness and its ability to effectively process and analyze high-dimensional data without prior simplification.

\paragraph{Impact of Dimensionality Reduction}
Interestingly, the performance of the Anomaly-Transformer model is positively influenced by dimensionality reduction techniques. Notably, when the data is reduced to half its dimensions (27, 12, 25) using PCA and Random Projection, the model maintains a high level of accuracy, particularly in terms of F1-score, for the MSL dataset. This suggests that the model is capable of adapting to reduced-dimensional data while retaining its anomaly detection accuracy.

\paragraph{Best Overall Performance}
The most notable performance of the Anomaly-Transformer model is observed with Random Projection reducing the dataset to 3 dimensions for the MSL dataset. This particular scenario highlights the model's flexibility and its ability to maintain high anomaly detection accuracy even when working with significantly reduced data dimensions.

\subsection{Dimensionality Reduction Techniques}\label{subsec2}

Dimensionality reduction techniques play a pivotal role in preprocessing high-dimensional data, particularly in the context of anomaly detection in time series. This section evaluates the effectiveness of various dimensionality reduction techniques - PCA, UMAP, Random Projection, and t-SNE - in enhancing the performance of the MUTANT and Anomaly-Transformer models. By comparing these techniques, we aim to identify their individual and comparative strengths and limitations. The focus is not only on how these techniques influence the accuracy and efficiency of the models but also on their impact on the interpretability of the results. This analysis is crucial for understanding the practical applicability of each dimensionality reduction technique in real-world scenarios.

\subsubsection{PCA}\label{subsubsec1}
Across the datasets, PCA demonstrates a consistent ability to maintain high model performance. The technique, particularly when reducing data to the lowest dimensions, shows notable improvements in F1-scores for the SMAP and SWaT datasets, indicating its effectiveness in simplifying data while preserving essential features for anomaly detection.

\subsubsection{UMAP}\label{subsubsec2}
UMAP distinguishes itself in enhancing the performance of the MUTANT model, especially for the SWaT dataset. The improvement in F1-score with UMAP reduction to the lowest dimension highlights its capability in effectively managing complex industrial datasets, making it a valuable tool in the dimensionality reduction arsenal.

\subsubsection{Rand. Proj.}\label{subsubsec3}
Random Projection, particularly when reducing the dataset to 3 dimensions, significantly enhances the performance of the Anomaly-Transformer model for the MSL dataset. This result underscores the technique's utility in lower-dimensional spaces, offering a balance between data simplification and preservation of anomaly detection capabilities.

\subsubsection{t-SNE}\label{subsubsec4}
t-SNE proves to be particularly effective for the SMAP dataset in the context of the Anomaly-Transformer model. This technique's focus on preserving local structures seems to resonate well with the dataset's inherent characteristics, leading to substantial performance enhancements, especially in terms of F1-score.

\subsection{Datasets}\label{subsec3}

The choice of dataset in anomaly detection research significantly influences the performance and applicability of models and dimensionality reduction techniques. In this section, we scrutinize the performance of the MUTANT and Anomaly-Transformer models across three distinct datasets - SMAP, MSL, and SWaT. Each dataset presents its own set of challenges and characteristics, offering a diverse range of scenarios to test the models' capabilities. This comparative analysis across different datasets aims to provide insights into the versatility and adaptability of the models under study, highlighting how different data characteristics can impact the effectiveness of anomaly detection approaches.

\subsubsection{SMAP}\label{subsubsec1}
The SMAP dataset benefits from dimensionality reduction, with techniques like PCA and t-SNE showing significant improvements in model performances, particularly in terms of F1-scores. This suggests that these techniques are well-suited for handling the specific characteristics and challenges presented by the SMAP dataset.

\subsubsection{MSL}\label{subsubsec2}
For the MSL dataset, both Random Projection and UMAP prove to be effective, with UMAP showing the best overall performance in enhancing the MUTANT model's F1-score. This indicates the adaptability of these techniques to the dataset's unique structure and anomaly detection requirements.

\subsubsection{SWaT}\label{subsubsec3}
The SWaT dataset sees its highest performance improvements with UMAP reduction in the MUTANT model, particularly in terms of F1-score. This highlights UMAP's effectiveness in dealing with the complexities and nuances of industrial control system data, as represented in the SWaT dataset.

\subsection{General Observations}\label{subsec4}

Through the course of this study, several key observations have emerged that offer critical insights into the interplay between dimensionality reduction techniques and anomaly detection models. Firstly, it is evident that dimensionality reduction typically enhances the performance of anomaly detection models. Techniques such as UMAP and Random Projection have shown particular promise, consistently improving model performance across various datasets.

A crucial aspect that has become apparent is the importance of the alignment between the chosen dimensionality reduction technique, the dataset at hand, and the specific anomaly detection model being used. This alignment is instrumental in achieving optimal performance, as different datasets and models respond uniquely to various dimensionality reduction methods.

Notably, the MUTANT model has demonstrated a pronounced affinity for UMAP reduction, particularly when applied to industrial datasets like SWaT, where it excels in handling complex data structures. On the other hand, the Anomaly-Transformer model has displayed remarkable versatility, adapting effectively across a range of dimensionality reduction techniques without compromising on its anomaly detection capabilities.

\begin{center}
    \captionof{table}{Comparison of Model Training Times Across Various Dimensionalities for MUTANT and Anomaly-Transformer on SWaT, SMAP, and MSL Datasets}
    \label{tab:dataset_training_time_details}
    \begin{tabular}{ccccc} 
        \toprule
        Model & Dataset & Dimensionality & CPU/GPU & \begin{tabular}[c]{@{}c@{}}Usage Time\\(Hours)\end{tabular}  \\ 
        \midrule

        \multirow{9}{*}{\Centering{}{\rotatebox[origin=c]{90}{MUTANT}}} & SWaT & Original & CPU & 65.15 \\
        & SWaT    & 25             & CPU     & 20.07               \\
        & SWaT    & 8              & CPU     & 9.62                \\
        \cdashline{2-5}[1pt/1pt]
        & SMAP    & Original       & CPU     & 4.67                \\
        & SMAP    & 12             & CPU     & 3.91                \\
        & SMAP    & 8              & CPU     & 3.08                \\
        \cdashline{2-5}[1pt/1pt]                                      
        & MSL     & Original       & CPU     & 6.34                \\
        & MSL     & 27             & CPU     & 2.61                \\
        & MSL     & 8              & CPU     & 1.24                \\
        \hline
        \multirow{12}{*}{\Centering{}{\rotatebox[origin=c]{90}{Anomaly-Transformer}}} & SWaT & Original & GPU & 1.49 \\
        & SWaT    & 25             & GPU     & 1.49                \\
        & SWaT    & 3              & GPU     & 1.50                \\
        & SWaT    & 2              & GPU     & 1.49                \\
        \cdashline{2-5}[1pt/1pt]
        & SMAP    & Original       & GPU     & 0.58                \\
        & SMAP    & 12             & GPU     & 0.58                \\
        & SMAP    & 3              & GPU     & 0.58                \\
        & SMAP    & 2              & GPU     & 0.58                \\
        \cdashline{2-5}[1pt/1pt]
        & MSL     & Original       & GPU     & 0.54                \\
        & MSL     & 27             & GPU     & 0.54                \\
        & MSL     & 3              & GPU     & 0.54                \\
        & MSL     & 2              & GPU     & 0.54                \\

        \bottomrule
    \end{tabular}
\end{center}

\begin{center}
    \includegraphics[width=5in]{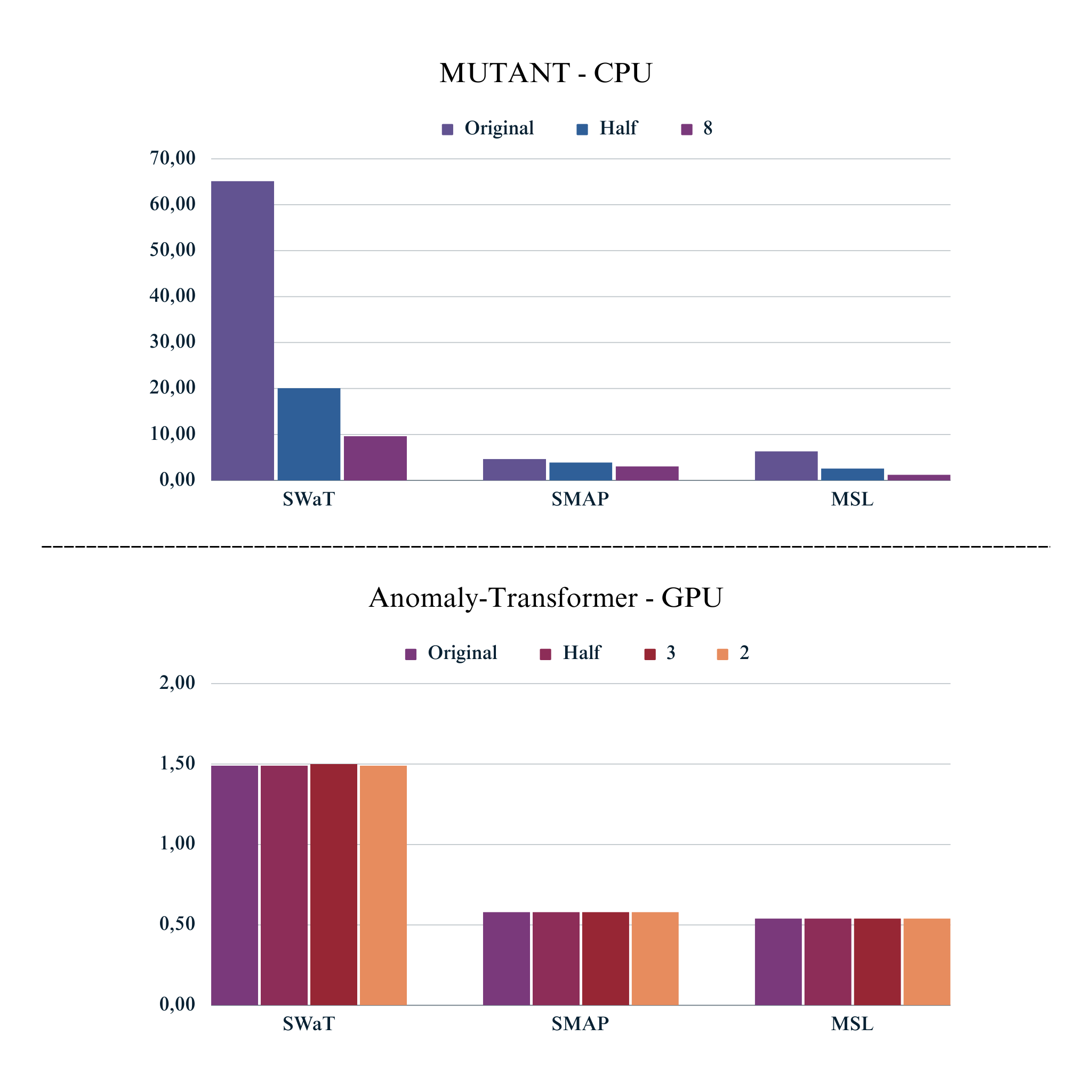} 
    \captionof{figure}{Comparison of Model Training Times Across Various Dimensionalities for MUTANT and Anomaly-Transformer on SWaT, SMAP, and MSL Datasets}
    \label{fig:dataset_training_time_details}
\end{center}

Additionally, our findings extend to the realm of operational efficiency, specifically in terms of model training times. The data presented in Table \ref{tab:dataset_training_time_details} and in Figure \ref{fig:dataset_training_time_details} illustrates a significant reduction in training times with the application of dimensionality reduction techniques. For instance, training the MUTANT model on the original SWaT dataset using CPU took 65.15 hours, while reducing the dimensionality to 25 and 8 resulted in training times of 20.07 and 9.62 hours, respectively. This represents a dramatic reduction in training time, particularly when the dimensionality is minimized to 8, underscoring the efficiency gains achievable through dimensionality reduction. Similar trends are observed with other datasets like SMAP and MSL, where reduced dimensions led to a marked decrease in training duration.

In contrast, the Anomaly-Transformer model exhibited consistent training times across various levels of dimensionality reduction when trained on a GPU, as seen with the SWaT, SMAP, and MSL datasets. This consistency in training duration, despite changes in dimensionality, highlights the model's robustness and the efficiency of GPU utilization. These findings collectively emphasize the dual benefit of dimensionality reduction: enhancing model accuracy and efficiency while also significantly reducing computational resource utilization and training time. Such efficiency is particularly crucial in scenarios involving large datasets or where rapid deployment of models is essential.

\section{Conclusion and Future Work}\label{sec6}

\subsection{Conclusion}\label{subsec1}
This study embarked on a comprehensive evaluation of two advanced unsupervised time series anomaly detection models, MUTANT and Anomaly-Transformer, across three distinct datasets — MSL, SMAP, and SWaT — and under different dimensionality reduction scenarios. Our findings reveal the nuanced interactions between the models, the dimensionality reduction techniques, and the dataset characteristics, offering valuable insights into their collective impact on anomaly detection.

The MUTANT model, with its unique architecture, showed notable adaptability, particularly when paired with the UMAP dimensionality reduction technique. This combination was especially effective in handling the complex data structures of the industrial SWaT dataset, demonstrating the model's capability in diverse settings. On the other hand, the Anomaly-Transformer model displayed a versatile nature, exhibiting robust performance across various dimensionality reduction techniques. Random Projection, in particular, emerged as a technique that maintained high performance levels for this model, highlighting its adaptability.

One of the most significant findings of this study is the role of dimensionality reduction techniques not only in simplifying data but also in enhancing the anomaly detection capabilities of the models. Techniques such as PCA, UMAP, and Random Projection struck a balance between reducing data complexity and preserving critical features necessary for effective anomaly detection.

Additionally, our empirical results extend to the realm of operational efficiency, specifically in the context of training times. By halving the dimensionality of data, we observed an average reduction in training times by approximately 300\%. More strikingly, minimizing the data to its lowest dimensions led to an even greater reduction in training times, averaging around 650\%. This substantial decrease in the time required to train models highlights the practical benefits of dimensionality reduction in scenarios where time and computational resources are limited. It underscores the importance of dimensionality reduction not just as a means to enhance model accuracy but also as a strategy to improve the overall efficiency of the anomaly detection process.

\subsection{Future Work}\label{subsec2}
Looking forward, the field of anomaly detection in time series data presents several avenues for further research and development:
\begin{itemize}
  \item \textbf{Exploration of Additional Datasets:} Testing these models on a wider array of datasets, including those from different domains, would provide deeper insights into their generalizability and adaptability.
  \item \textbf{Hybrid Anomaly Detection Approaches:} Investigating hybrid models that combine various anomaly detection methods could lead to more robust and accurate solutions.
  \item \textbf{Real-Time Anomaly Detection:} Adapting these models for real-time anomaly detection in streaming data remains a significant challenge, offering substantial practical value, particularly in industrial and IoT applications.
  \item \textbf{Interpretable AI in Anomaly Detection:} Enhancing the interpretability of these models, especially in the context of their decision-making processes, would increase their trustworthiness and applicability in critical decision-making scenarios.
  \item \textbf{Advanced Dimensionality Reduction Techniques:} Exploring newer or more sophisticated dimensionality reduction methods could further improve the performance of anomaly detection models.
\end{itemize}

In conclusion, this study not only reinforces the importance of model and technique selection based on specific dataset characteristics but also opens new perspectives on the potential of dimensionality reduction in enhancing anomaly detection models. The continuous evolution of techniques and models in this field promises further advancements, contributing to more efficient and effective solutions in time series anomaly detection.

\section*{Declarations}

\backmatter
\bmhead{Data and Code Availability} The dataset and source code pertinent to this study can be accessed at the following URL: \url{https://github.com/mahsunaltin/DR4MTSAD}.
\bmhead{Funding} This study is supported by the TUBITAK 121F065 and ITU-BAP Doctoral Projects 2024, which are conducted by Altan Cakir.


\bibliography{sn-bibliography}

\end{document}